\documentclass[conference]{IEEEtran}
\IEEEoverridecommandlockouts

\usepackage[hyphens]{url}
\usepackage[hidelinks]{hyperref}
\hypersetup{breaklinks=true}
\usepackage{cite}
\usepackage{amsmath,amssymb,amsfonts}
\usepackage{algorithmic}
\usepackage{graphicx}
\usepackage{subcaption}
\usepackage{textcomp}
\usepackage{xcolor}
\usepackage{tabularx,booktabs}
\newcolumntype{Y}{>{\centering\arraybackslash}X}

\def\BibTeX{{\rm B\kern-.05em{\sc i\kern-.025em b}\kern-.08em
    T\kern-.1667em\lower.7ex\hbox{E}\kern-.125emX}}

\begin{document}

\title{FACTUAL: A Novel \underline{F}r\underline{a}mework for \underline{C}on\underline{t}rastive Learning Based Rob\underline{u}st S\underline{A}R Image C\underline{l}assification
\thanks{
Accepted by the 2024 IEEE Radar Conference.

© 2024 IEEE. Personal use of this material is permitted. Permission from IEEE must be
obtained for all other uses, in any current or future media, including
reprinting/republishing this material for advertising or promotional purposes, creating new
collective works, for resale or redistribution to servers or lists, or reuse of any copyrighted
component of this work in other works.}}

\author{\IEEEauthorblockN{Xu Wang\IEEEauthorrefmark{1},
Tian Ye\IEEEauthorrefmark{1},
Rajgopal Kannan\IEEEauthorrefmark{2}, 
Viktor Prasanna\IEEEauthorrefmark{1}}
\IEEEauthorblockA{\IEEEauthorrefmark{1,2,4}University of Southern California \IEEEauthorrefmark{2}DEVCOM Army Research Lab}
\IEEEauthorblockA{\IEEEauthorrefmark{1,2,4}\{xwang032, tye69227, prasanna\}@usc.edu \IEEEauthorrefmark{2} rajgopal.kannan.civ@army.mil}
}

\maketitle

\begin{abstract}
Deep Learning (DL) Models for Synthetic Aperture Radar (SAR) Automatic Target Recognition (ATR), while delivering improved performance, have been shown to be quite vulnerable to adversarial attacks.
Existing works improve robustness by training models on adversarial samples. However, by focusing mostly on attacks that manipulate images randomly, they neglect the real-world feasibility of such attacks. 
In this paper, we propose FACTUAL, a novel 
Contrastive Learning framework for Adversarial Training and robust
SAR classification.
FACTUAL consists of two  components: (1) Differing from existing works, a novel perturbation scheme that incorporates realistic physical adversarial attacks (such as OTSA) to build a supervised adversarial pre-training network. This network utilizes class labels for clustering clean and perturbed images together into a more informative feature space. (2) A linear classifier  cascaded after the encoder to use the computed representations to predict the target labels. 
By pre-training and fine-tuning our model on both clean and adversarial samples, we show that our model achieves high prediction accuracy on both cases. Our model achieves 99.7\% accuracy on clean samples, and 89.6\% on perturbed samples, both outperforming previous state-of-the-art methods.
\end{abstract}


\section{Introduction}


 Synthetic Aperture Radar (SAR) based remote sensing systems are widely used in both military and civilian fields due to their all-weather and all-time (day-and-night) operability\cite{7572958}. Concomitantly, there is an increasing need for automated exploitation of SAR image processing tasks such as Automatic Target Recognition (ATR) which aims to find and predict the class of targets within SAR images. Recently, Deep Learning (DL) models have demonstrated great performance on SAR ATR. By extracting useful latent information to compute representations for SAR images, they have demostrated impressive accuracy in downstream classification tasks.

Nonetheless, the susceptibility of DL models remains a severe security concern, especially in SAR image classification, given its critical implications in the military domain. Adversarial samples, generated by renowned attacks like the Fast Gradient Sign Method (FGSM)\cite{DBLP:journals/corr/GoodfellowSS14} and Projected Gradient Descent (PGD)\cite{DBLP:conf/iclr/MadryMSTV18}, have almost indistinguishable variations from the original image, which can lead to misclassifications by neural networks. Such attacks exploit the model's weak points by taking advantage of the inconsistencies and roughness of the model’s feature space. As a countermeasure, Adversarial Training (AT)\cite{DBLP:journals/corr/GoodfellowSS14} has been devised to bolster a model’s resilience by training it using samples produced from various attack strategies. 

Interestingly, the concept of AT dovetails seamlessly with Contrastive Learning principles. Recent studies have discovered the effectiveness of applying Contrastive Learning (CL) methods in robust representation learning for computer vision tasks. Moreover, its application has been expanded to include SAR image classification. Contrastive Learning method couples an image with its augmented versions as positive pairs, and contrasts this image with randomly chosen ones as negative pairs. The objective is to minimize the distance between the positive pairs while maximizing the distance between the negative pairs in the feature space. This ensures a more refined feature space within the model, which makes the model more robust against adversarial attacks. In \cite{DBLP:conf/nips/KimTH20 }\cite{DBLP:conf/nips/JiangCCW20}, Adversarial Contrastive Learning has been proposed, which constructs positive pairs using original images and the corresponding perturbed images that maximize the contrastive loss. In \cite{DBLP:journals/remotesensing/XuSCLJK21}, an Unsupervised Adversarial Contrastive Learning (UACL) framework is proposed for SAR image. 
Previous works on Adversarial Contrastive Learning suffer from two major drawbacks: (1) They do not utilize class label information during pre-training. Consequently, pre-trained models only cluster instance-wise positive pairs together, while losing information about the similarity of other images belonging to the same class. (2) They generate adversarial samples without considering the physical feasibility of the attacks on SAR images. Therefore, such pre-trained models may not be robust against real-world physical attacks. 

In this paper, we propose FACTUAL, a novel Supervised Contrastive Learning (SCL)\cite{DBLP:conf/nips/KhoslaTWSTIMLK20} based framework for robust SAR  classification that (1) develops more {\it informative} and {\it effective}  model pre-training by utilizing class-label information to  cluster clean and perturbed images  together and (2) achieves {\it higher classification robustness} to realistic physical attacks such as the On-Target Scatterer Attack (OTSA)\cite{2312.02912}. Specifically, we develop a novel Adversarial CL component in FACTUAL that targets  perturbations to only the object region in SAR images. By utilizing such adversarial samples together with PGD for Contrastive Learning, our model learns more informative representations for superior robustness. 


We summarize our contributions below:

\begin{itemize}
    \item We generate adversarial samples using both PGD and OTSA. By training models using clean samples together with adversarial examples generated by two adversaries, we show the multi-modality of different nature of adversarial attacks on SAR images efficiently exploits model's weakness, and improves model's robustness against attacks.
    \item We introduce SCL into the process of adversarial pre-training for SAR image classification. We show that utilizing label information significantly improve model robustness.
    \item We show that our method achieves 99.7\% accuracy on clean samples and 89.6\% accuracy on perturbed samples, both outperforming the previous state-of-the-art methods. 
\end{itemize}

\section{Background}

\subsection{Adversarial Attacks}

Initially introduced by \cite{DBLP:journals/corr/SzegedyZSBEGF13}, deep neural networks learn input-output mappings that are highly discontinuous. As a result, even slight perturbations that are imperceptible can mislead the neural networks to make wrong predictions. Inspired from that, \cite{DBLP:journals/corr/GoodfellowSS14} proposed Fast Gradient Sign Method (FGSM). By perturbing the input along the direction of the gradient of the loss function, FGSM efficiently exploits the discontinuity of the feature space, and generates perturbed samples that are indistinguishable to the original samples, yet leading to wrong predictions. Specifically, let $\theta$ be the parameters of the model, $x$ be the input, $y$ be its label and $J(\theta, x, y)$ be the loss function. FGSM generates a constrained perturbation $\eta$ by computing

\begin{equation}
    \eta = \epsilon\cdot\text{sign}(\nabla_x J(\theta,x,y)),
\end{equation}
where $\epsilon$ is a small positive constant that constrains the magnitude of perturbation. Such attack can be interpreted as a one-step attack that tries to maximize the loss. To be more powerful, \cite{DBLP:conf/iclr/MadryMSTV18} proposed the PGD attack, which is a multi-step variant of FGSM. PGD searches in the constrained space iteratively to find the perturbation that maximize the loss. Let $S$ be the search space of perturbations. Let $x$ be the data at $0$ iteration. At the $t+1$ iteration, PGD perturbs the sample by computing

\begin{equation}
    x^{t+1} = \Pi_{x+S} (x^t + \epsilon\cdot\text{sign}(\nabla_x J(\theta,x^t,y))),
\end{equation}

Where $\Pi$ is the projection function that finds the point in $x+S$ that is closest to $x^t + \epsilon\cdot\text{sign}(\nabla_x J(\theta,x^t,y))$. By iteratively searching in the constrained space and perturbing the sample, PGD finds perturbations that maximize the loss to a greater extent, compared to FGSM. There are also more gradient-based attacks proposed in the later research. Gradient-based attacks are always considered one of the most commonly use attacks, since they generate adversarial samples along the fastest direction that increases the loss, and therefore are capable of exploiting model's weak points and successfully result in misclassification. 


Although gradient-based attacks have shown a superior ability to decrease a deep learning model's performance, it is non-trivial to apply them directly on SAR images, due to the unique imagery mechanism of SAR. Since a SAR imagery system is usually planted on the moving platform internally such as a plane, it is hard to directly add perturbations on SAR images. Therefore, \cite{2312.02912} proposed the On-Target Scatter Attack (OTSA) which takes into consideration the feasibility of executing the attack on SAR images in the real world. It executes the attack by physically attaching carefully designed scatterers to the on-ground object. 
As a gradient-based attack, OTSA initialized its perturbations on a fixed number of pixels that are in the region of the target in the image. While iteratively searching in the constrained space to find more potent perturbations in a gradient-based attack manner, it enforces that the perturbations are made in the target region of the image. This feasible approach enables rich future applications in attacking images of moving vehicles. 

\begin{figure*}[t]
\centerline{\includegraphics[scale=1.4]{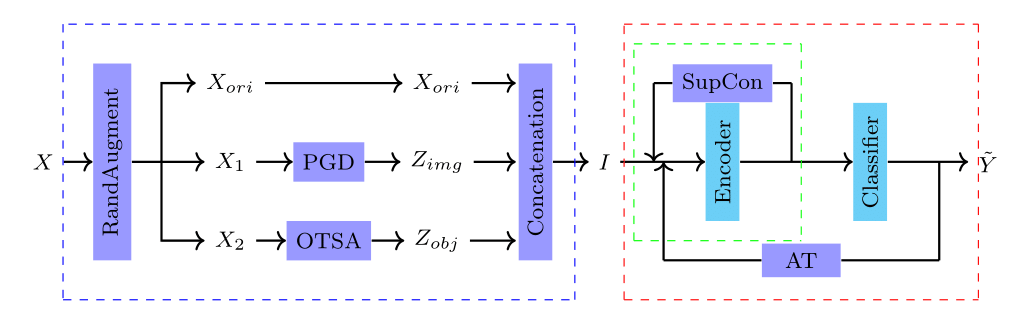}}
\caption{Framework architecture of FACTUAL. The region with blue dashed line as border refers to the Data Augmentation step. The region with green dashed line as border refers to the Supervised Adversarial Contrastive Pre-training step. The region with Red dashed line as border refers to the Supervised Adversarial Fine-tuning step. }
\label{model}
\end{figure*}

\subsection{Contrastive Learning}

Recent studies have discovered the importance of the Contrastive Learning (CL) method in boosting model performance on computer vision tasks. The CL method, as initially introduced in techniques such as MoCo\cite{DBLP:conf/cvpr/He0WXG20} and SimCLR\cite{DBLP:conf/icml/ChenK0H20}, operates as a self-supervised representation learning approach. It pulls an anchor image together with the positive sample in the feature space, and pushes away the negative sample from the anchor. Since labels are usually unavailable in the setting of self-supervised learning, a positive sample of the anchor is often generated by data augmentations, and a negative sample is the sample that is randomly selected from the batch of the data. CL ensures a refined feature space where representations of positive pairs are clustered together, and is especially effective on improving model's performance on downstream tasks when the feasible dataset is of relative small size. 

However, initial methodologies generally operate on the assumption that random samples are unlikely to be from the same class. To address the challenge of false negative pairs, SCL\cite{DBLP:conf/nips/KhoslaTWSTIMLK20} was introduced. By integrating labeled data, SCL only considers negative samples from classes distinct from the original image. Meanwhile, Bootstrap Your Own Latent (BYOL)\cite{DBLP:conf/nips/GrillSATRBDPGAP20} championed the use of solely positive pairs for Contrastive Learning, significantly cutting down the requisite batch size for training. Another proposition \cite{DBLP:conf/cvpr/Yuan0K0WMKF21} is to harness the multi-modality of input data. In \cite{DBLP:journals/corr/abs-2209-02329}, data collected from different sensors at the same locations are treated as a natural way of data augmentation, which is then modeled using CL and greatly enriches the information space. Representations derived from these techniques have proven to be superior in subsequent tasks, such as image classification. In light of this, recent studies have advocated for reducing the distance between clean and perturbed input during the pre-training stage. \cite{DBLP:conf/nips/KimTH20 } proposed Robust Contrastive Learning (RoCL) which generates adversarial samples for Contrastive Learning without using any labels. In \cite{DBLP:conf/nips/JiangCCW20}, the thorough discussion of variants of Adversarial Contrastive Learning (ACL) concludes that using both clean and perturbed samples for Contrastive Learning could result in the state-of-the-art trade-off between model's performance on clean and perturbed samples. Later in \cite{DBLP:journals/remotesensing/XuSCLJK21}, an Unsupervised Adversarial Contrastive Learning (UACL) framework is proposed specifically for SAR image. As pointed out by \cite{DBLP:conf/cvpr/Chen0C0AW20}\cite{DBLP:conf/nips/HendrycksMKS19}, a model trained with CL achieves better robustness, and fine-tuning the pre-trained model improves the robustness even further. 

Compared with previous works on Adversarial Contrastive Learning on SAR images, our work pushes the limit from both sides. From the perspective of Adversarial Training, our work incorporates the most commonly-use adversarial attack on general images and the physically executable attacks specifically designed for SAR. From the perspective of Contrastive Learning, our work utilizes label information to further refine the feature space. Our contributions on both sides maximize the use of information contained in clean samples and perturbed samples, which results in superior performance on downstream classfication tasks. 

\section{Method}

In this section, our proposed Adversarial Contrastive Learning framework is described in detail. We first introduce the overall framework and then explain each component specifically.

\subsection{Overall Framework}

Our overall framework consists of three parts: data augmentation, supervised adversarial contrastive pre-training, and supervised adversarial fine-tuning. As shown in Fig.~\ref{model}, we first augment our data using standard augmentations and adversarial attacks. We then pass our augmented data to the encoder, and optimize the encoder using supervised contrastive loss. A projector will be applied on top of the encoder during the pre-training phase, and will be discarded after that. During the fine-tuning, we cascade a linear classifier after the encoder. Given images' representations obtained from the encoder, our linear classifier will predict the class of objects in the images. The full network will then be optimized iteratively by minimizing the cross entropy loss.

\subsection{Data Augmentation}
Denote the original dataset by $I_{ori}$ with $N$ images. Given each original image $x\in I_{ori}$, we first randomly augment it into two samples: $(x_1, x_2)$. Each data augmentation could be viewed as a different view of the original image, which will help improve the model's ability of generalization when the dataset is small. We then use OTSA to generate adversarial perturbations targeting on the object in the image, and use PGD to generate perturbations on the whole image, denoting them by $\delta_{obj}$ and $\delta_{img}$, respectively. PGD and OTSA represent two distinct types of distributions of perturbations on the image. Pre-training our model with these two attacks could improve the model’s generalizability and robustness to unseen attacks. We then apply the perturbations to the augmented images and obtain perturbed samples $z_{img}=x_1+\delta_{img}$ and $z_{obj}=x_2+\delta_{obj}$, respectively. As pointed out by \cite{DBLP:conf/nips/JiangCCW20}, if we train the model using only perturbed samples, the model will achieve the strongest robustness, which means that its prediction accuracy on perturbed samples is the highest. However, the model will be overftting to perturbed samples and as a result, its performance on clean samples degrades. To balance this trade-off while maximizing the use of perturbed samples, we enhance Contrastive Learning with three types of images. Combined with the original image, we obtain a triple $(x, z_{obj}, z_{img})$ for each original image. Apply the same procedure to all the original images, we obtain our augmented dataset $I$, where the size of the dataset is $|I| = 3N$.

\begin{figure*}[tb]

\begin{subfigure}{0.33\textwidth}
\includegraphics[width=0.9\linewidth, height=5cm]{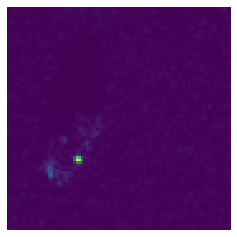} 
\caption{Sample SAR image}
\label{fig:subim1}
\end{subfigure}
\begin{subfigure}{0.33\textwidth}
\includegraphics[width=0.9\linewidth, height=5cm]{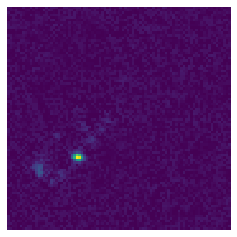}
\caption{Sample perturbed by PGD}
\label{fig:subim2}
\end{subfigure}
\begin{subfigure}{0.33\textwidth}
\includegraphics[width=0.9\linewidth, height=5cm]{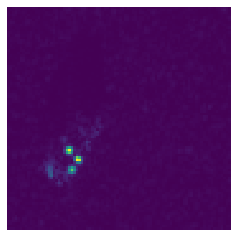}
\caption{Sample perturbed by OTSA}
\label{fig:subim3}
\end{subfigure}
\caption{Clean sample and perturbed samples of SAR image}
\label{fig:sar}
\end{figure*}

\subsection{Supervised Adversarial Contrastive Pre-training}

Initialized randomly, we use common deep learning models as our encoder. Given augmented dataset $I$, we aim to train an encoder $f$ which maps images into the representation space, in which images that belong to the same class are clustered together, while different clusters are distributed way from each other. Let $i \in I$ be any given sample, $\theta_{enc}$ be the parameters of our encoder, we define

\begin{equation}
    f_i = f(\theta_{enc}, i),
\end{equation}

We then pre-train the encoder by combining SCL with AT, which is introduced by \cite{DBLP:conf/iclr/MadryMSTV18}. Let $x$ denote a given clean sample, $D$ denote the training set, $\delta$ denote the perturbation, $S$ denote the set of feasible perturbations, $L$ denote the loss function, $\theta$ denote the parameters of the model and loss function, and $y$ denote the label. Adversarial Training solves the following min-max problem

\begin{equation}
    \min_{\theta} \mathbb{E}_{x \sim D} \left[ \max_{\delta \in S} L(\theta, x + \delta, y) \right],
\end{equation}

In our method, we aim to firstly minimize the Supervised Contrastive Loss over our augmented dataset. Let $I = I_{ori} \cup I_{obj} \cup I_{img}$ be the set of representations of samples, $i$ be an arbitrary sample, $f_i$ be its representation, $A(i)\equiv I\setminus \{i\}$, and $P(i)$ be the set of all samples that belong to the same class as $i$.  Formally, our adversarial contrastive loss is given by
\begin{equation}
     L_{cl, i} =  \frac{-1}{|P(i)|}\sum_{p\in P(i)} log\frac{exp(f_i \cdot f_p/\tau)}{\sum_{a\in A(i)} exp(f_i \cdot f_a / \tau)},
\end{equation}
where $\tau$ is a scalar temperature parameter.  We then rewrite the objective of Supervised Contrastive Loss as 

 \begin{equation}
     \min_{\theta_{enc}} \mathbb{E}_{i \sim I} [L_{cl,i}(\theta_{enc}, i)],
 \end{equation}

\subsection{Supervised Adversarial Fine-tuning}

After pre-training, we cascade a Linear Classifier (LC) on top of the encoder and use the obtained representations to make predictions on the image labels. Let $H = f(I)$ be the set of output representations and $Y$ be the set of ground-truth labels. We forward the representations to a linear classifier and then compute the cross entropy loss. Let $\theta_{dis}$ be the parameters of each linear classifier. Let $\theta = \{\theta_{enc}, \theta_{dis}\}$, we optimize the full network using the cross entropy loss:

\begin{equation}
\theta = \text{argmin}_{\theta}L_{ce}(LC(H), Y),
\end{equation}

\section{Experiment}
\subsection{Dataset}

Moving and Stationary Target Acquisition and Recognition (MSTAR) dataset is a standard dataset produced by US Defense Resarch Project Agency\cite{10.1117/12.242059}. It consists of high-resolution SAR images of 10 classes of military vehicles including 2S1, BTR70, BMP2, BRDM2, BTR60, D7, T62, T72, ZIL131 and ZSU23. The training set contains images that are collected at depression angle 17°, and the test set contains images collected at 15°. Each image contains only one target. The MSTAR dataset is commonly used to test the SAR ATR algorithms. Since the size of the dataset is relatively small and the number of images that belong to each class is not equal, \cite{7460942} proposed to pre-process the dataset by randomly sampling a number of  88 $\times$ 88 patches from the original 128 $\times$ 128 images. Following the procedures in \cite{7460942}, the augmented MSTAR dataset contains 27,000 images for training, and 2,425 images for testing. In the training set, each class contains about 2,700 images.

After augmenting the MSTAR following the precedures in \cite{7460942} from the level of dataset, we further use RandAugment\cite{DBLP:conf/nips/CubukZS020} to augment the dataset from the level of image. RandAugment is a collection of augmentations which can be drawn randomly during the training phase of deep learning models. In our experiment, it consists of following transforms: Resize, Crop, Flip, Color, Brightness, Contrast, Saturation, and Hue. Augmenting dataset creates different views of the original image, which helps improve model's generalizability.

\subsection{Attacks}
We use OTSA to generate adversarial samples on targets in the SAR image and use PGD to generate adversarial samples on the whole image. For OTSA, we generate 3 scatters for each input image and restrict them to be in the region of the targets of the image. We then run 10 iterations to update the scatters in the region of the target. For PGD, we set the maximum perturbation $\epsilon$ as 8/255, the maximum number of iterations to generate adversarial samples as 7. For other hyperparameters of OTSA and PGD, we follow their default settings\cite{2312.02912}\cite{DBLP:conf/iclr/MadryMSTV18}. An example of clean and perturbed SAR images is shown in Fig.~\ref{fig:sar}. Combined with the original image, the size of the final training dataset is $|I| = 81,000$. 

\subsection{Model}

There are a variety of deep learning models proposed for encoding images. For illustration purposes, we use the ResNet50 as our encoder to test the effectiveness of our method. ResNet50 is a variant of ResNet\cite{DBLP:conf/cvpr/HeZRS16}, which is one of the most widely used deep learning models in computer vision tasks. However, we point out that our method can be extended to any deep learning models. 

For the classifier, we append a simple linear classifier to the end of each encoder. The linear classifier only consists of one-step linear mapping that maps each representation to a prediction vector with dimension equals number of classes. 

\subsection{Metrics}
Following \cite{DBLP:conf/cvpr/Chen0C0AW20}, we report three metrics:
\begin{itemize}
    \item Standard Test Accuracy (TA): model's prediction accuracy on clean sample. 
    \item Robust Test Accuracy (RA): model's prediction accuracy on perturbed images generated from previously seen attacks.
    \item Average Accuracy (AA) : model's prediction accuracy on all samples. 
\end{itemize}
As pointed out in \cite{DBLP:conf/iclr/MadryMSTV18}, there exists a trade-off between accuracy and adversarial robustness during Adversarial Training. In the real world, models being extremely robust against perturbed samples while performing badly on clean samples cannot be employed. 
Therefore, it is important to report TA and RA separately to better evaluate the model’s robustness against adversarial attacks without degrading classification performance on clean samples. 
Besides, we also report the gap between RA and TA. Achieving a relatively small gap indicates that the model is not overfitted to one type of samples, which implies strong generalizability. 

\subsection{Experiment Design}

We first evaluate our fine-tuned models on unseen clean and perturbed samples and report the TA and RA. Our perturbed samples consist of adversarial samples generated by PGD and OTSA. The RA is computed as the accuracy on all unseen perturbed samples.
For comparison, we train a ResNet50 model using only clean samples, referred to as Standard Training (ST).

Besides, we demonstrate that utilizing label information during pre-training will significantly enhance the effectiveness of Adversarial Contrastive Learning and benefit the downstream classification on MSTAR. As \cite{DBLP:journals/remotesensing/XuSCLJK21} systematically benchmarked different defense methods against adversarial attacks on MSTAR, we follow exactly the same settings as  \cite{DBLP:journals/remotesensing/XuSCLJK21}, and list our results with theirs in Table~\ref{tab2}. We also show the performance of standard AT for reference purpose. As our methods and the baseline methods all use ResNet as the encoder while baseline methods are in a self-supervised learning manner, this comparison will clearly demonstrate the effectiveness of our method. As other methods' performance on OTSA attacks are unavailable, we only compare our method's performance with other methods on their performance against PGD attacks.

\begin{table}[tb]
\caption{Evaluation of ResNet-50 trained with different methods against PGD and OTSA}
\begin{center}
\begin{tabularx}{0.5\textwidth}{c *{5}{Y}}
\toprule
Method
 & \multicolumn{4}{c}{Metrics}  \\
\cmidrule(lr){2-5} 
  & AA & TA & RA & TA-RA \\
\midrule
 ST \cite{DBLP:conf/cvpr/HeZRS16}    & 73.4  & 99.6 & 44.5 & 55.1\\
 FACTUAL & \textbf{96.1}  & \textbf{99.7} & \textbf{94.4} & \textbf{5.3}\\
\bottomrule
\end{tabularx}
\label{tab1}
\end{center}
\end{table}

\begin{table}[tb]
\caption{Comparison of different methods against PGD Attack}
\begin{center}
\begin{tabularx}{0.5\textwidth}{c *{5}{Y}}
\toprule
Metrics
 & \multicolumn{1}{c}{ResNet50} & \multicolumn{3}{c}{ResNet18} \\
\cmidrule(lr){3-6} 
        & FACTUAL   & AT\cite{DBLP:journals/corr/GoodfellowSS14}        & RoCL\cite{DBLP:conf/nips/KimTH20}    & ACL\cite{DBLP:conf/nips/JiangCCW20}       &  UACL\cite{DBLP:journals/remotesensing/XuSCLJK21}\\
\midrule
 TA     & \textbf{99.7} & 60.47 &  92.43  &  95.34    &  95.09   \\
 RA     & \textbf{89.6} &  60.47 & 80.73  &  34.43    &  80.92   \\
 TA-RA & \textbf{10.1} & 0 & 11.7  & 60.9 & 14.1\\
\bottomrule
\end{tabularx}
\label{tab2}
\end{center}
\end{table}

\subsection{Experiment Results}
Table~\ref{tab1} shows the comparison between FACTUAL and ST.
It can be observed that FACTUAL improves RA by 50\% while maintaining the highest TA of 99.7\%, which indicates that FACTUAL does well on preventing itself from being overfitted to perturbed samples. From Table~\ref{tab2}, FACTUAL yields a significant improvement on model's robustness against PGD attacks. Compared with previous state-of-the-art methods, FACTUAL achieves a 99.7\% prediction accuracy on clean samples, and a 89.6\% prediction accuracy on perturbed samples. Besides, FACTUAL achieves an impressive small gap between its TA and RA. While the gap between RA and TA is significant for all the previous state-of-the-art methods and even the minimum gap is 11.7\% reached by RoCL, FACTUAL outperforms all of them by achieving a 10.1\% gap. With superior performance on both clean and perturbed samples while maintaining the smallest gap between RA and TA, FACTUAL demonstrates its effectiveness on learning more informative representations which significantly benefits the downstream tasks. 

\section{Conclusion}

In this paper, we addressed the problem of building robust deep learning models for SAR ATR. We proposed FACTUAL, the novel Contrastive Learning based Adversarial Training framework, that not only utilize the multimodality of different natures of adversarial attacks on SAR images, but also leverages label information for better representation learning. By testing our method's impressive performance on unseen perturbed samples as well as clean samples, we demonstrate that our method could learn informative representations that yield superior accuracy and robustness for classification tasks. 

\section*{Acknowledgment}

This work is supported by the DEVCOM Army Research
Lab (ARL) under grant W911NF2320186.
\\

\textbf{Distribution Statement A}: Approved for public release.
Distribution is unlimited.

\bibliographystyle{IEEEtran}
\bibliography{main}

\begin{thebibliography}{10}
\providecommand{\url}[1]{#1}
\csname url@samestyle\endcsname
\providecommand{\newblock}{\relax}
\providecommand{\bibinfo}[2]{#2}
\providecommand{\BIBentrySTDinterwordspacing}{\spaceskip=0pt\relax}
\providecommand{\BIBentryALTinterwordstretchfactor}{4}
\providecommand{\BIBentryALTinterwordspacing}{\spaceskip=\fontdimen2\font plus
\BIBentryALTinterwordstretchfactor\fontdimen3\font minus \fontdimen4\font\relax}
\providecommand{\BIBforeignlanguage}[2]{{%
\expandafter\ifx\csname l@#1\endcsname\relax
\typeout{** WARNING: IEEEtran.bst: No hyphenation pattern has been}%
\typeout{** loaded for the language `#1'. Using the pattern for}%
\typeout{** the default language instead.}%
\else
\language=\csname l@#1\endcsname
\fi
#2}}
\providecommand{\BIBdecl}{\relax}
\BIBdecl

\bibitem{7572958}
K.~El-Darymli, E.~W. Gill, P.~Mcguire, D.~Power, and C.~Moloney, ``Automatic target recognition in synthetic aperture radar imagery: A state-of-the-art review,'' \emph{IEEE Access}, vol.~4, pp. 6014--6058, 2016.

\bibitem{DBLP:journals/corr/GoodfellowSS14}
\BIBentryALTinterwordspacing
I.~J. Goodfellow, J.~Shlens, and C.~Szegedy, ``Explaining and harnessing adversarial examples,'' in \emph{3rd International Conference on Learning Representations, {ICLR} 2015, San Diego, CA, USA, May 7-9, 2015, Conference Track Proceedings}, Y.~Bengio and Y.~LeCun, Eds., 2015. [Online]. Available: \url{http://arxiv.org/abs/1412.6572}
\BIBentrySTDinterwordspacing

\bibitem{DBLP:conf/iclr/MadryMSTV18}
\BIBentryALTinterwordspacing
A.~Madry, A.~Makelov, L.~Schmidt, D.~Tsipras, and A.~Vladu, ``Towards deep learning models resistant to adversarial attacks,'' in \emph{6th International Conference on Learning Representations, {ICLR} 2018, Vancouver, BC, Canada, April 30 - May 3, 2018, Conference Track Proceedings}.\hskip 1em plus 0.5em minus 0.4em\relax OpenReview.net, 2018. [Online]. Available: \url{https://openreview.net/forum?id=rJzIBfZAb}
\BIBentrySTDinterwordspacing

\bibitem{DBLP:conf/nips/KimTH20}
\BIBentryALTinterwordspacing
M.~Kim, J.~Tack, and S.~J. Hwang, ``Adversarial self-supervised contrastive learning,'' in \emph{Advances in Neural Information Processing Systems 33: Annual Conference on Neural Information Processing Systems 2020, NeurIPS 2020, December 6-12, 2020, virtual}, H.~Larochelle, M.~Ranzato, R.~Hadsell, M.~Balcan, and H.~Lin, Eds., 2020. [Online]. Available: \url{https://proceedings.neurips.cc/paper/2020/hash/1f1baa5b8edac74eb4eaa329f14a0361-Abstract.html}
\BIBentrySTDinterwordspacing

\bibitem{DBLP:conf/nips/JiangCCW20}
\BIBentryALTinterwordspacing
Z.~Jiang, T.~Chen, T.~Chen, and Z.~Wang, ``Robust pre-training by adversarial contrastive learning,'' in \emph{Advances in Neural Information Processing Systems 33: Annual Conference on Neural Information Processing Systems 2020, NeurIPS 2020, December 6-12, 2020, virtual}, H.~Larochelle, M.~Ranzato, R.~Hadsell, M.~Balcan, and H.~Lin, Eds., 2020. [Online]. Available: \url{https://proceedings.neurips.cc/paper/2020/hash/ba7e36c43aff315c00ec2b8625e3b719-Abstract.html}
\BIBentrySTDinterwordspacing

\bibitem{DBLP:journals/remotesensing/XuSCLJK21}
\BIBentryALTinterwordspacing
Y.~Xu, H.~Sun, J.~Chen, L.~Lei, K.~Ji, and G.~Kuang, ``Adversarial self-supervised learning for robust {SAR} target recognition,'' \emph{Remote. Sens.}, vol.~13, no.~20, p. 4158, 2021. [Online]. Available: \url{https://doi.org/10.3390/rs13204158}
\BIBentrySTDinterwordspacing

\bibitem{DBLP:conf/nips/KhoslaTWSTIMLK20}
\BIBentryALTinterwordspacing
P.~Khosla, P.~Teterwak, C.~Wang, A.~Sarna, Y.~Tian, P.~Isola, A.~Maschinot, C.~Liu, and D.~Krishnan, ``Supervised contrastive learning,'' in \emph{Advances in Neural Information Processing Systems 33: Annual Conference on Neural Information Processing Systems 2020, NeurIPS 2020, December 6-12, 2020, virtual}, H.~Larochelle, M.~Ranzato, R.~Hadsell, M.~Balcan, and H.~Lin, Eds., 2020. [Online]. Available: \url{https://proceedings.neurips.cc/paper/2020/hash/d89a66c7c80a29b1bdbab0f2a1a94af8-Abstract.html}
\BIBentrySTDinterwordspacing

\bibitem{2312.02912}
T.~Ye, R.~Kannan, V.~Prasanna, C.~Busart, and L.~Kaplan, ``Realistic scatterer based adversarial attacks on sar image classifiers,'' in \emph{2023 IEEE International Radar Conference (RADAR)}, 2023, pp. 1--6.

\bibitem{DBLP:journals/corr/SzegedyZSBEGF13}
\BIBentryALTinterwordspacing
C.~Szegedy, W.~Zaremba, I.~Sutskever, J.~Bruna, D.~Erhan, I.~J. Goodfellow, and R.~Fergus, ``Intriguing properties of neural networks,'' in \emph{2nd International Conference on Learning Representations, {ICLR} 2014, Banff, AB, Canada, April 14-16, 2014, Conference Track Proceedings}, Y.~Bengio and Y.~LeCun, Eds., 2014. [Online]. Available: \url{http://arxiv.org/abs/1312.6199}
\BIBentrySTDinterwordspacing

\bibitem{DBLP:conf/cvpr/He0WXG20}
\BIBentryALTinterwordspacing
K.~He, H.~Fan, Y.~Wu, S.~Xie, and R.~B. Girshick, ``Momentum contrast for unsupervised visual representation learning,'' in \emph{2020 {IEEE/CVF} Conference on Computer Vision and Pattern Recognition, {CVPR} 2020, Seattle, WA, USA, June 13-19, 2020}.\hskip 1em plus 0.5em minus 0.4em\relax Computer Vision Foundation / {IEEE}, 2020, pp. 9726--9735. [Online]. Available: \url{https://doi.org/10.1109/CVPR42600.2020.00975}
\BIBentrySTDinterwordspacing

\bibitem{DBLP:conf/icml/ChenK0H20}
\BIBentryALTinterwordspacing
T.~Chen, S.~Kornblith, M.~Norouzi, and G.~E. Hinton, ``A simple framework for contrastive learning of visual representations,'' in \emph{Proceedings of the 37th International Conference on Machine Learning, {ICML} 2020, 13-18 July 2020, Virtual Event}, ser. Proceedings of Machine Learning Research, vol. 119.\hskip 1em plus 0.5em minus 0.4em\relax {PMLR}, 2020, pp. 1597--1607. [Online]. Available: \url{http://proceedings.mlr.press/v119/chen20j.html}
\BIBentrySTDinterwordspacing

\bibitem{DBLP:conf/nips/GrillSATRBDPGAP20}
\BIBentryALTinterwordspacing
J.~Grill, F.~Strub, F.~Altch{\'{e}}, C.~Tallec, P.~H. Richemond, E.~Buchatskaya, C.~Doersch, B.~{\'{A}}. Pires, Z.~Guo, M.~G. Azar, B.~Piot, K.~Kavukcuoglu, R.~Munos, and M.~Valko, ``Bootstrap your own latent - {A} new approach to self-supervised learning,'' in \emph{Advances in Neural Information Processing Systems 33: Annual Conference on Neural Information Processing Systems 2020, NeurIPS 2020, December 6-12, 2020, virtual}, H.~Larochelle, M.~Ranzato, R.~Hadsell, M.~Balcan, and H.~Lin, Eds., 2020. [Online]. Available: \url{https://proceedings.neurips.cc/paper/2020/hash/f3ada80d5c4ee70142b17b8192b2958e-Abstract.html}
\BIBentrySTDinterwordspacing

\bibitem{DBLP:conf/cvpr/Yuan0K0WMKF21}
\BIBentryALTinterwordspacing
X.~Yuan, Z.~Lin, J.~Kuen, J.~Zhang, Y.~Wang, M.~Maire, A.~Kale, and B.~Faieta, ``Multimodal contrastive training for visual representation learning,'' in \emph{{IEEE} Conference on Computer Vision and Pattern Recognition, {CVPR} 2021, virtual, June 19-25, 2021}.\hskip 1em plus 0.5em minus 0.4em\relax Computer Vision Foundation / {IEEE}, 2021, pp. 6995--7004. [Online]. Available: \url{https://openaccess.thecvf.com/content/CVPR2021/html/Yuan\_Multimodal\_Contrastive\_Training\_for\_Visual\_Representation\_Learning\_CVPR\_2021\_paper.html}
\BIBentrySTDinterwordspacing

\bibitem{DBLP:journals/corr/abs-2209-02329}
\BIBentryALTinterwordspacing
U.~Jain, A.~Wilson, and V.~Gulshan, ``Multimodal contrastive learning for remote sensing tasks,'' \emph{CoRR}, vol. abs/2209.02329, 2022. [Online]. Available: \url{https://doi.org/10.48550/arXiv.2209.02329}
\BIBentrySTDinterwordspacing

\bibitem{DBLP:conf/cvpr/Chen0C0AW20}
\BIBentryALTinterwordspacing
T.~Chen, S.~Liu, S.~Chang, Y.~Cheng, L.~Amini, and Z.~Wang, ``Adversarial robustness: From self-supervised pre-training to fine-tuning,'' in \emph{2020 {IEEE/CVF} Conference on Computer Vision and Pattern Recognition, {CVPR} 2020, Seattle, WA, USA, June 13-19, 2020}.\hskip 1em plus 0.5em minus 0.4em\relax Computer Vision Foundation / {IEEE}, 2020, pp. 696--705. [Online]. Available: \url{https://openaccess.thecvf.com/content\_CVPR\_2020/html/Chen\_Adversarial\_Robustness\_From\_Self-Supervised\_Pre-Training\_to\_Fine-Tuning\_CVPR\_2020\_paper.html}
\BIBentrySTDinterwordspacing

\bibitem{DBLP:conf/nips/HendrycksMKS19}
\BIBentryALTinterwordspacing
D.~Hendrycks, M.~Mazeika, S.~Kadavath, and D.~Song, ``Using self-supervised learning can improve model robustness and uncertainty,'' in \emph{Advances in Neural Information Processing Systems 32: Annual Conference on Neural Information Processing Systems 2019, NeurIPS 2019, December 8-14, 2019, Vancouver, BC, Canada}, H.~M. Wallach, H.~Larochelle, A.~Beygelzimer, F.~d'Alch{\'{e}}{-}Buc, E.~B. Fox, and R.~Garnett, Eds., 2019, pp. 15\,637--15\,648. [Online]. Available: \url{https://proceedings.neurips.cc/paper/2019/hash/a2b15837edac15df90721968986f7f8e-Abstract.html}
\BIBentrySTDinterwordspacing

\bibitem{10.1117/12.242059}
\BIBentryALTinterwordspacing
E.~R. Keydel, S.~W. Lee, and J.~T. Moore, ``{MSTAR extended operating conditions: a tutorial},'' in \emph{Algorithms for Synthetic Aperture Radar Imagery III}, E.~G. Zelnio and R.~J. Douglass, Eds., vol. 2757, International Society for Optics and Photonics.\hskip 1em plus 0.5em minus 0.4em\relax SPIE, 1996, pp. 228 -- 242. [Online]. Available: \url{https://doi.org/10.1117/12.242059}
\BIBentrySTDinterwordspacing

\bibitem{7460942}
S.~Chen, H.~Wang, F.~Xu, and Y.-Q. Jin, ``Target classification using the deep convolutional networks for sar images,'' \emph{IEEE Transactions on Geoscience and Remote Sensing}, vol.~54, no.~8, pp. 4806--4817, 2016.

\bibitem{DBLP:conf/nips/CubukZS020}
\BIBentryALTinterwordspacing
E.~D. Cubuk, B.~Zoph, J.~Shlens, and Q.~Le, ``Randaugment: Practical automated data augmentation with a reduced search space,'' in \emph{Advances in Neural Information Processing Systems 33: Annual Conference on Neural Information Processing Systems 2020, NeurIPS 2020, December 6-12, 2020, virtual}, H.~Larochelle, M.~Ranzato, R.~Hadsell, M.~Balcan, and H.~Lin, Eds., 2020. [Online]. Available: \url{https://proceedings.neurips.cc/paper/2020/hash/d85b63ef0ccb114d0a3bb7b7d808028f-Abstract.html}
\BIBentrySTDinterwordspacing

\bibitem{DBLP:conf/cvpr/HeZRS16}
\BIBentryALTinterwordspacing
K.~He, X.~Zhang, S.~Ren, and J.~Sun, ``Deep residual learning for image recognition,'' in \emph{2016 {IEEE} Conference on Computer Vision and Pattern Recognition, {CVPR} 2016, Las Vegas, NV, USA, June 27-30, 2016}.\hskip 1em plus 0.5em minus 0.4em\relax {IEEE} Computer Society, 2016, pp. 770--778. [Online]. Available: \url{https://doi.org/10.1109/CVPR.2016.90}
\BIBentrySTDinterwordspacing

\end{thebibliography}

\end{document}